\documentclass{article} 
\usepackage{arxiv,times}

\usepackage{amsmath,amsfonts,bm}

\def\eqref#1{equation~\ref{#1}}

\def\1{\bm{1}}

\DeclareMathAlphabet{\mathsfit}{\encodingdefault}{\sfdefault}{m}{sl}
\SetMathAlphabet{\mathsfit}{bold}{\encodingdefault}{\sfdefault}{bx}{n}

\usepackage{hyperref}
\usepackage{url}           
\usepackage{booktabs}      
\usepackage{amsfonts}      
\usepackage{nicefrac}       
\usepackage{microtype}      
\usepackage{xcolor}        
\usepackage{amsmath}
\usepackage{amssymb}
\usepackage{graphicx}
\usepackage{natbib}
\usepackage{booktabs}
\usepackage{tabularx}
\usepackage{xcolor}
\usepackage{adjustbox}
\usepackage{xspace}
\usepackage{colortbl} 
\usepackage{soul} 
\usepackage{tabularx}
\usepackage{stfloats}
\usepackage{algorithm}
\usepackage[noend]{algpseudocode}
\usepackage{algcompatible}
\usepackage{wrapfig}
\usepackage{pifont}
\usepackage{makecell}
\graphicspath{{./figures/}}
\usepackage{multirow}

\definecolor{lightgrey}{HTML}{dcdbdb}
\sethlcolor{lightgrey}
\definecolor{lightblue}{HTML}{E8F0FE}
\definecolor{lightblue}{HTML}{E8F0FE}
\definecolor{gray}{HTML}{9aa0a6}
\definecolor{lightpink}{HTML}{F48FB1}
\definecolor{lightred}{HTML}{FFCBC9}
\definecolor{lightcyan}{HTML}{80DEEA}
\definecolor{lightgrey}{HTML}{dcdbdb}
\sethlcolor{lightgrey}

\newcommand{\cc}[0]{\cellcolor{lightblue}}

\newcommand{\ModelName}{Personalized Visual Instruction Tuning\xspace}
\newcommand{\ModelNameAbbre}{PVIT\xspace}
\newcommand{\Bench}{P-Bench\xspace}
\usepackage{tcolorbox}
\newcommand{\Data}{PVIT-3M\xspace}
\newtcolorbox[auto counter, number within=section, list type=subsubsection, list inside=toc]{sectionbox}[2][]{
colback=white!98!gray, colframe=black, 
colbacktitle=white!90!gray, coltitle=black, 
fonttitle=\bfseries,
title={#2}, 
list entry={Comment \thetcbcounter\quad}
}

\title{\raisebox{-0.35cm}{\includegraphics[width=1.5cm]{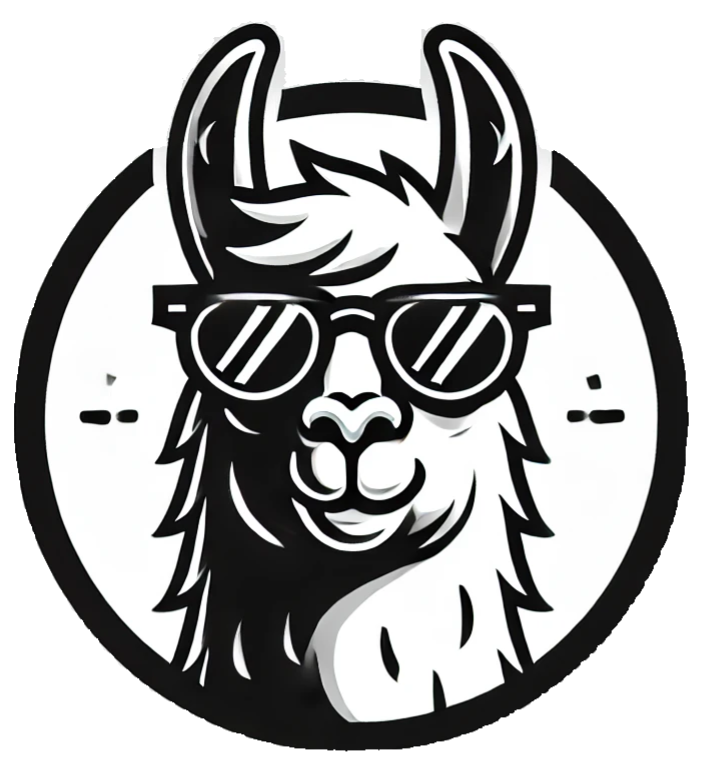}} \ModelName}

\author{Renjie Pi$^1$\footnotemark[1], \quad Jianshu Zhang$^{1}$\thanks{\, Equal Contribution.\\Code and data are available at the following links: \\\url{https://github.com/sterzhang/PVIT}\\ \url{https://huggingface.co/datasets/Sterzhang/PVIT-3M}. \\The code and data are released under MIT and apache2.0 licenses, respectively.}, 
\quad\textbf{Tianyang Han}$^1$,
\quad\textbf{Jipeng Zhang}$^1$,\quad Rui Pan$^2$,\quad
\textbf{Tong Zhang$^2$}
\\
  $^1$The Hong Kong University of Science and Technology\quad
$^2$University of Illinois Urbana-Champaign \\
\texttt{\{rpi,jzhanggr\}@ust.hk,} \quad\texttt{jianshu.zhang@whu.edu.cn,} \\\texttt{23104841g@connect.polyu.hk,} \quad \texttt{ruip4@illinois.edu,}\\
\texttt{tongzhang@tongzhang-ml.org}
}

\finalcopy 
\begin{document}

\maketitle
\begin{abstract}
Recent advancements in multimodal large language models (MLLMs) have demonstrated significant progress; however, these models exhibit a notable limitation, which we refer to as ``face blindness". Specifically, they can engage in general conversations but fail to conduct personalized dialogues targeting at specific individuals. This deficiency hinders the application of MLLMs in personalized settings, such as tailored visual assistants on mobile devices, or domestic robots that need to recognize members of the family. In this paper, we introduce \textbf{\ModelName (\ModelNameAbbre)}, a novel data curation and training framework designed to enable MLLMs to identify target individuals within an image and engage in personalized and coherent dialogues. Our approach involves the development of a sophisticated pipeline that autonomously generates training data containing personalized conversations. This pipeline leverages the capabilities of various visual experts, image generation models, and (multi-modal) large language models. To evaluate the personalized potential of MLLMs, we present a benchmark called \Bench, which encompasses various question types with different levels of difficulty. The experiments
demonstrate a substantial personalized performance enhancement after fine-tuning with our curated dataset.
\end{abstract}
\section{Introduction}
The advent of Large Language Models (LLMs) has significantly advanced AI, transforming natural language processing and understanding~\citep{openlm2023openllama, openai2023gpt4, touvron2023llama, scao2022bloom, chowdhery2022palm, alpaca, vicuna2023}. These models, trained on extensive text corpora, possess substantial world knowledge, excelling in various tasks. Progress in LLMs has led to rapid enhancements in Multimodal Large Language Models (MLLMs)~\citep{liu2023llava, zhu2023minigpt4, dai2023instructblip, li2023blip2, openai2023gpt4, bai2023qwenvl}. MLLMs leverage pretrained visual encoders (e.g., vision transformers) to process images, incorporating them as token embeddings alongside text token embeddings. These models expand LLM capabilities to engage in conversations based on image inputs, offering diverse applications like autonomous driving~\citep{ding2023hilm} and medical assistants~\citep{li2023llavamed}.

Despite the success of MLLMs, their effectiveness is limited in general purpose conversations, and drastically fail at personalized conversations targeting at specific individuals. For example, given a photograph of a girl named Lisa, and an image with Lisa inside of it, the state-of-the-art MLLMs are not able to recognize her and provide corresponding information. This deficiency prohibits the use of the MLLMs in personalized use cases, such as your personal AI assistent deployed on the mobile phone, a domestic robot that needs to recognize and serve your family members, or a smart home system that requires individualized interactions to cater to the specific needs of different residents.

To empower MLLMs with personalization capability (denoted  as P-MLLM), previous endeavors propose to augment the MLLM with external heads and vocabularies, which are trained to identify specific individuals within a scene using a few personalized training data~\citep{alaluf2024myvlmpersonalizingvlmsuserspecific, nguyen2024yollavapersonalizedlanguagevision}. Although these approaches demonstrate good performances, they suffer from the following weaknesses: 1) they require additional training for each newly introduced individual, which is inflexible and unpractical for real-life scenarios, since the person of interest may frequently change; 2) it can not be guaranteed that we can always collect the training data for the person of interest. Therefore, it is undoubtedly more promising if the personalization capability of the MLLM is able to generalize, rather than being limited on a few pre-defined individuals.

Owing to the auto-regressive training paradigm adopted for current state-of-the-art LLMs and MLLMs, they possess the ability to generate responses depending on a given prefix. This ability is also referred to as in-context learning capability~\citep{wei2023chainofthoughtpromptingelicitsreasoning}, which enables the model to produce different outputs during inference by adjusting the prefix without further training. Therefore, one intuitive and practical approach is to provide the information of individuals to the MLLM in its prefix. In this way, the MLLM is expected to provide answers for different input individuals during inference. However, our experimental results indicate that this capability is challenging, and current MLLMs struggle to effectively comprehend such personalized inputs. This may be due to the fact that these MLLMs are fine-tuned with limited multimodal data and a lack of personalized data, which together hinder their ability to develop in-context understanding for multimodal inputs.

To address this challenge, we propose \textbf{\ModelName (\ModelNameAbbre)}, a novel training paradigm that enables MLLMs to perceive personalized inputs as in-context prefixes. Specifically, each individual is represented as a $<$personal image, personal introduction$>$ pair, which is provided to the MLLM as a \textit{multi-modal prefix}. We further introduce \textit{personalized wrapper tokens} to group the visual and textual information of each individual, thereby eliminating ambiguity when multiple individuals are involved. During training, the MLLM is optimized to answer questions related to target individuals within the prefixes. Once trained, the MLLM is able to fully utilize its in-context learning capability, and generalizes to arbitrary individuals without requiring additional fine-tuning or modifications to the model architecture.

In our proposed \ModelNameAbbre paradigm, the most critical barrier is the absence of large-scale, high-quality training data. 
To address this challenge, we design an automatic framework to synthesize personalized training data, operating in three phases. First, visual expert models extract individual visual concepts from scene images (\textit{Visual Concept Curation}). Next, we utilize the MLLM to convert these visual concepts into both individual-level and scene-level textual descriptions, which are fused to create a coherent representation (\textit{Dual-Level Textual Information Extraction and Fusion}). Finally, LLMs generate diverse personalized QA pairs using reasoning and instruction-following capabilities (\textit{PVIT Dataset Generation}).
To evaluate the personalization capability of MLLMs, we further created a benchmark termed \textbf{\Bench}, which assesses personalization capability from multiple perspectives. The results indicate that the ability of current SOTA MLLMs to perceive personalized concepts is limited, which can be significantly boosted after training with our proposed \ModelNameAbbre.

To summarize, we make the following contributions in this paper:
\begin{itemize}
    \item Inspired by in-context learning capabilities of LLMs, we propose \textbf{\ModelName (\ModelNameAbbre)}, a new training paradigm that equips MLLMs to conduct personalized conversation for any arbitrary individuals with no addition training at inference.
    \item We meticulously design an automatic data annotation framework to curate high-quality personalized training data, and synthesize a large scale dataset to enhance the MLLM's capability to conduct personalized conversations.
    \item We curate \textbf{\Bench}, a novel benchmark for evaluating the personalization capability of MLLMs. We demonstrate that the MLLM trained with our curated dataset demonstrates significantly improved performances.
\end{itemize}
\begin{figure}[t!]
\includegraphics[width=1.0\textwidth]{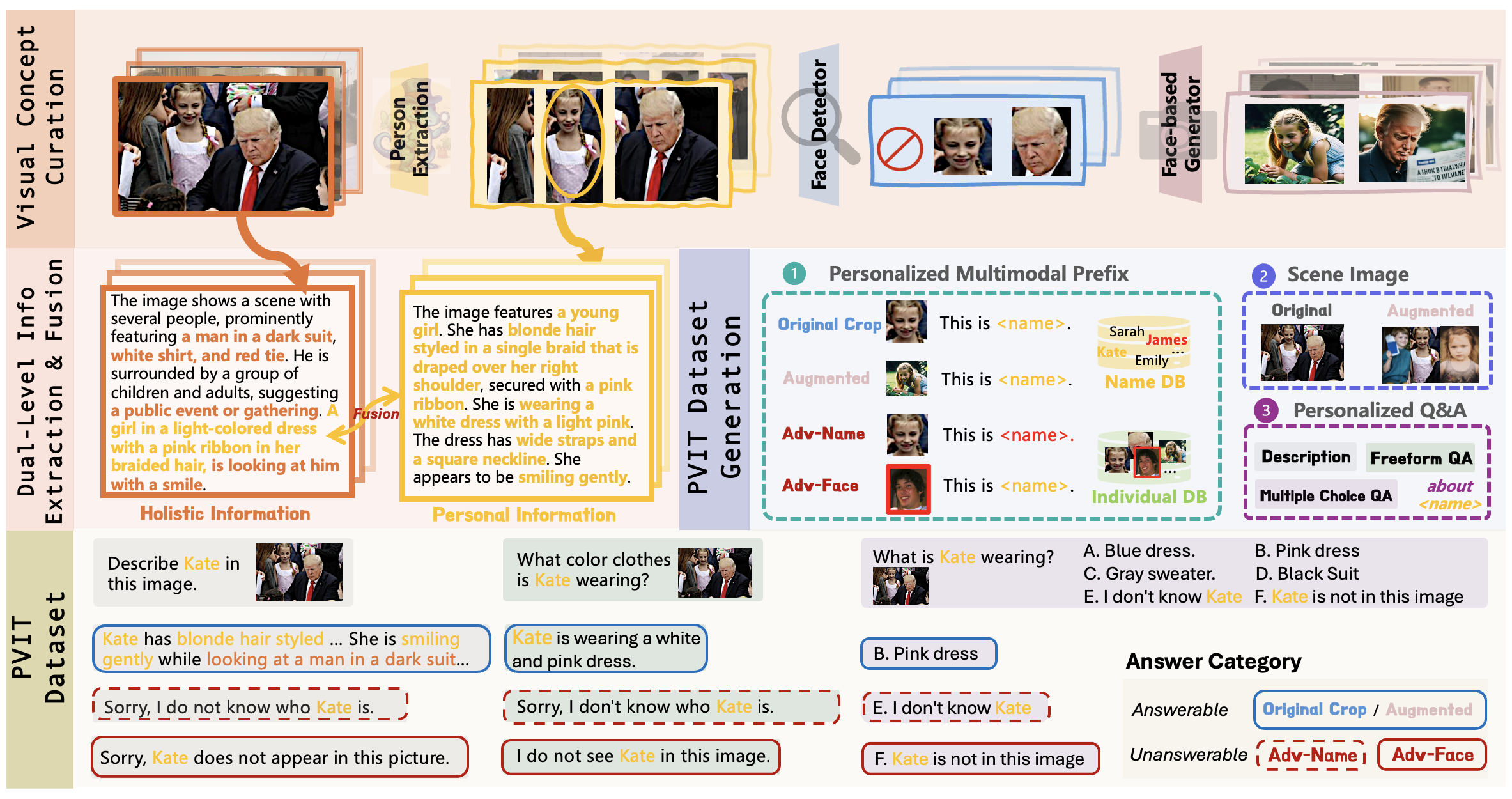} 
\vspace{-0.6cm}
\caption{ The \textbf{\ModelName \space (\ModelNameAbbre)} framework consists of three phases. In the \textit{visual concept curation} phase, we extract individuals and their faces from images, then augment them with different poses and angles. During the \textit{dual-level textual information extraction and fusion} phase, MLLMs first generate both holistic information and personal information, then integrate them to get more detailed and contextually accurate information. In the \textit{PVIT dataset generation} phase, LLMs create QA pair templates based on the extracted textual information, which are filled with selected names to construct training data.
}\label{fig:framework}
\vspace{-1.5em}
\end{figure}
\section{Related Work}

\paragraph{Multi-Modal Large Language Model.}
Recent advancements in large language models (LLMs) have significantly improved language comprehension and generation, achieving near-human proficiency across various tasks~\citep{brown2020language, scao2022bloom, chowdhery2022palm, smith2022using, hoffmann2022training, ouyang2022training, touvron2023llama, bai2022training}. This success has spurred interest in vision-language interaction, leading to multi-modal large language models (MLLMs)~\citep{liu2023llava, li2023blip2, dai2023instructblip, zhu2023minigpt4, dai2023instructblip, openai2023gpt4, bai2023qwenvl, su2023pandagpt, gao2023llamaadapter, pi2023detgpt, pi2023perceptiongpt, pi2024strengthening, pi2024mllmprotector, gao2023gllava}, which excel in dialogue based on visual inputs. However, they struggle with personalized conversations, such as dialogues about specific individuals like a girl named Kitty, limiting their use in personalized AI assistants, tailored recommendations, or customized therapy. Addressing this limitation is crucial for enabling personalized AI interactions.

\paragraph{Model Personalization}
In model personalization, the main objective is to tailor a model to learn new user-specific concepts. Considerable research has been conducted in the area of text-to-image (t2i) generation personalization~\citep{gal2023image,ruiz2022dreambooth,alaluf2023neural,arar2023domainagnostic,voynov2023p+,ye2023ip-adapter,wang2024instantid}. The predominant method for t2i personalization involves  fine-tuned the word embeddings using a few examples to capture the nuances of the target concept. Alternatively, some research has concentrated on personalizing image captioning models~\citep{chunseong2017attend,zeng2019automatic,wang2023user,park2018towards,zeng2019automatic,shuster2019engaging}. These personalized captioning methods aim to produce captions in a particular writing style. In contrast, our goal is to enable the model to integrate a new user-specific concept into a personalized textual description response that aligns with a given image. 

\paragraph{Personalized MLLMs} Compared to text-to-image generation and traditional image captioning, the personalization of Multimodal Large Language Models (MLLMs) remains an under-explored area. Recent approaches propose introducing new tunable parameters for each new individual and conduct tuning using a few training samples. Specifically, MyVLM~\citep{alaluf2024myvlmpersonalizingvlmsuserspecific} adopts a Q-former-style architecture, incorporating learnable heads to extract specific concepts and append them to the visual features. On the other hand, Yo'LLaVA~\citep{nguyen2024yollavapersonalizedlanguagevision} utilizes an LLaVA-style architecture, proposing to directly incorporate new concepts as additional tokens in the LLM's vocabulary. Despite these approaches demonstrating promising performance in personalized conversation, they require additional training and parameters for each newly introduced individual. This requirement makes the paradigm less practical in real life since new individuals are likely to be incorporated from time to time. In addition, it is often not possible to collect training samples associated with each individual. These limitations necessitate the model to generalize on new concepts on the fly.
\section{Problem Formulation}
Unlike previous methods for personalized multimodal large language models (P-MLLMs), which require adding trainable parameters and fine-tuning for each new individual, our approach treats personalization as an in-context learning task. This eliminates the need for fine-tuning on each new individual. More specifically, we first provide the list of \textbf{multi-modal prefixes} to the MLLM, which consists of $<$personal image, personal introduction$>$ pairs. Then, we present the \textbf{scene image} and a \textbf{personalized query} to conduct conversation targeting at specific individuals. Conditioned on the multi-modal prefixes and personalized query, the MLLM is expected to produce responses specific to the individual of interest accordingly.

\begin{figure}[t!]
\includegraphics[width=1.0\textwidth]{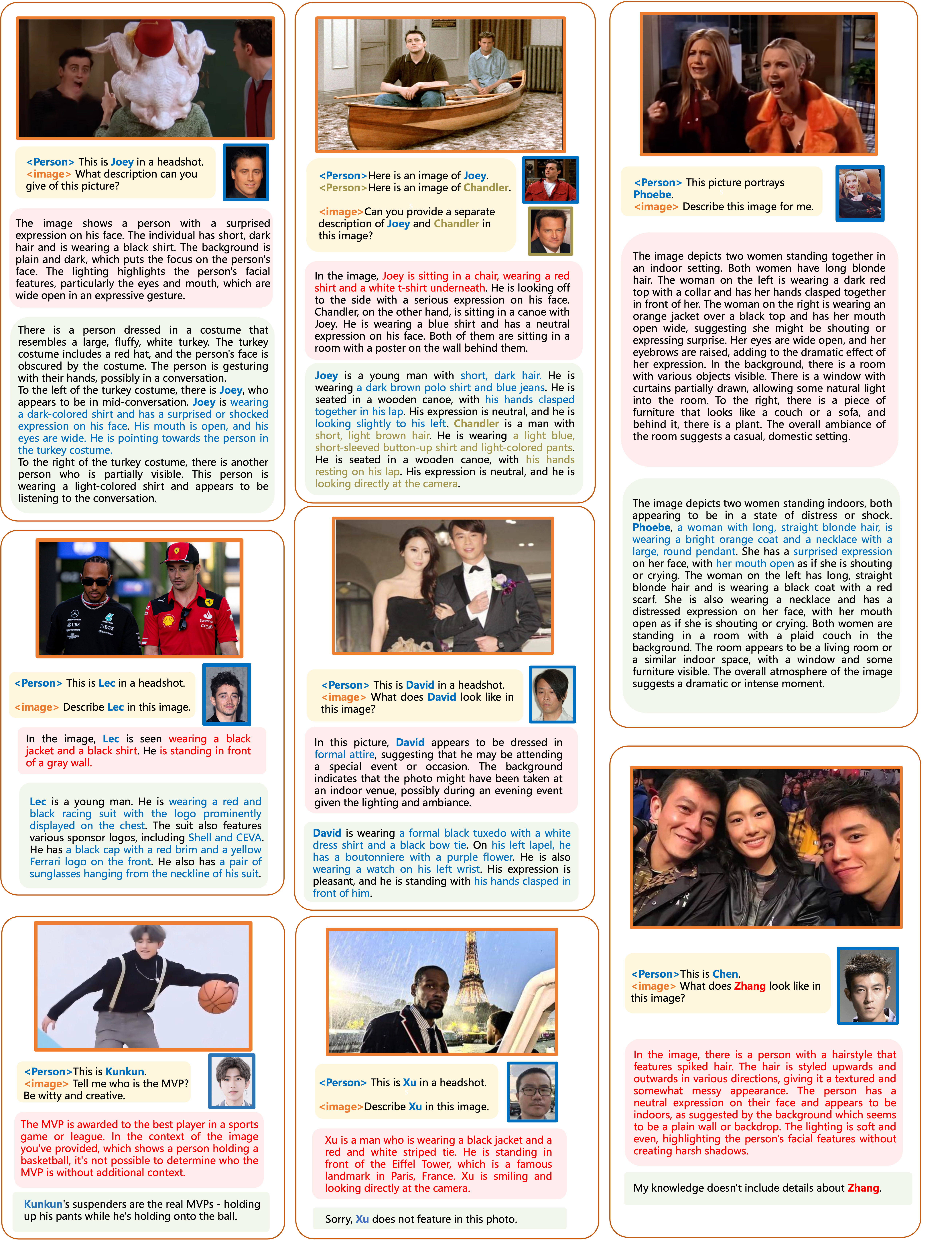} 
\vspace{-0.6cm}
\caption{Qualitative examples of P-LLaVA results: Each example includes the user's query, \textcolor{blue!70}{input individual photos}, and the \textcolor{orange}{scene image}. The \colorbox{red!10}{current MLLMs} fail to recognize the person of interest and conduct personalized conversations, whereas \colorbox{teal!8}{our model}, after training with \ModelNameAbbre, enables coherent and accurate personalized dialogues. Examples illustrate both answerable and unanswerable scenarios. For answerable cases, inputs involve single or multiple individuals, and our model incorporates names from the prefix for personalized responses. In unanswerable cases, current MLLMs provide incorrect answers, while the our model appropriately refuses and explains the reason.}

\label{fig:deom}
\end{figure}

We observe that this capability is nontrivial, which is not possessed by the majority of state-of-the-art MLLMs. As shown in Figure~\ref{fig:deom}, the MLLM tends to produce general responses, even when the user instruction is targeting at a specific individual. This issue persists even in MLLMs trained on interleaved image and text data. We presume that this limitation is due to the lack of training, rather than deficiency in the visual encoder.

Therefore, we propose to empower the MLLM with the personalization capability via \textbf{\ModelName} (\textbf{\ModelNameAbbre}). Specifically, we first collect a training dataset 
$\mathcal{D} = \left\{ \left( r^i, q^i, I^i_s, \left\{ I^i_{p_k}, t^i_{p_k} \right\}_{k=1}^{K^i} \right) \right\}_{i=1}^{N}
$
, where $r^i$, $q^i$ and $I^i_s$ stand for the $i^{th}$ target response, user query, and the scene image provided to the MLLM, respectively. On the other hand, $\left\{ I^i_{p_k}, t^i_{p_k} \right\}_{k=1}^{K^i}$ represent the $K^i$ multimodal prefixes containing images of input individuals and their personalized introduction, such as names. 
The optimization problem can be formulated as follows:
\begin{equation}
\begin{aligned}
    \mathcal{L}(\mathcal{D})&=-\frac{1}{N} \sum_{i=1}^{N}\sum^{L^i}_{s=1}\log p&\left[r^{i,s} | \mathcal{F} (r^{i,(<s)}, q^{i}, I_s^i, \left\{ I^i_{p_k}, t^i_{p_k} \right\}_{k=1}^{K^i})\right],
\end{aligned}
\end{equation}
During training, we minimize the auto-regressive loss to generate personalized responses based on the query, scene image, and multimodal prefixes. This approach fully harnesses the in-context learning capabilities of MLLMs, allowing them to adapt to new individuals without additional training.

\paragraph{Personalized Wrapper Tokens} Naively interleaving personalized images and introduction in the multimodal prefixes may introduce ambiguity, as personalized introduction (e.g., a name) could be mistakenly associated with either the preceding or following person's image. This ambiguity complicates training and confuses the MLLM during inference. To resolve this issue, we introduce two special tokens into the MLLM's vocabulary: $\langle\mid$ person\_start $\mid\rangle$. These tokens serve as wrappers to clearly enclose each individual’s information, structured as follows: $\langle\mid$ person\_start $\mid\rangle$ \{photo\} \{text\_intro\} $\langle\mid$ person\_end $\mid\rangle$. This design ensures that the information belonging to each individual is properly grouped, improving the model's learning process.

The main challenge of the in-context learning paradigm lies in the lack of personalized training data, as existing multimodal instruction-tuning datasets focus on general conversations. To address this, we propose an automated framework for annotating multimodal conversation data with personalized inputs, leveraging LLMs, MLLMs, image generation models, and vision experts. Details of this framework are discussed in the next section.
\section{Data Construction}
To equip the MLLM with personalized conversation capabilities, we developed a data generation framework that synthesizes various types of personalized training data~\citep{ye2022zerogen, meng2022generating, gao2023selfguided, pi2024image}. The framework begins by utilizing a set of images, which serve as scene images for the training data. It first extracts visual concepts of the individuals from these images using vision expert models. Then, it converts the visual information contained in the images into textual descriptions using MLLMs. Finally, it utilized the LLM's reasoning and instruction following capability to create high-quality and diverse QA  pairs. As illustrated in Figure~\ref{fig:framework}, the framework operates in three main phases: 1) Visual Concept Curation, 2) Dual-Level Textual Information Extraction and  Fusion, and 3) \ModelNameAbbre Dataset Generation. Below, we describe the procedure for each phase in detail.

\subsection{Visual Concept Curation}
In this phase, based on the scene images, we devise strategies for collecting images of individuals. First, we apply \textbf{person identification} to accurately locate the individuals within the images. Next, we perform \textbf{person augmentation} to generate images of the same individuals in various contexts and different perspectives. These images will later serve as the foundation for creating both visual and textual information of the input individuals, as well as personalized QA pairs.

\paragraph{Person Identification} For each image, we apply an open-vocabulary object detector, such as GroundingDino~\citep{liu2023grounding}, to localize individuals by providing the image and the text prompt ``$\texttt{a person}$". Next, for each detected person, we use a face detector~\citep{geitgey2016machine} to identify and localize the corresponding face, as the face is the most distinctive feature for identification. The images of the individuals along their faces are stored for later stages. Individuals without a detected face are excluded during this process.
\paragraph{Person Augmentation} 
After the previous step, we derive a list of $<$person, face$>$ pairs for each image. Each person in the scene image can be referenced by their corresponding face. However, in practice, the face used as reference to an individual often differs from the scene image. To introduce more variation in human faces and enhance the capability of MLLMs to recognize individuals, we adopt the identity preserving image generator PhotoMaker~\citep{li2023photomakercustomizingrealistichuman} to augment the person, which produces images of the same individual from different perspectives and contexts based on an input face. These augmented images can then be leveraged as reference to the individuals in the original scene images.

\subsection{Dual-Level Textual Information Extraction and  Fusion}

To construct personalized conversations for specific individuals in the subsequent phase, it is essential to not only derive the characteristics of each person in the scene image, but also capture how they are interacting with the surrounding context, which will serve as the basis for creating conversations that accurately reflect each individual's role and behavior in the scene image. To achieve this, we employ a  dual-level information extraction and fusion approach.

\paragraph{Personal Information Extraction} Since current MLLMs are unable to directly provide specific features of a designated person in the scene image containing more than one person, we focus on each individual by providing their cropped images, which are extracted from the previous phase, to the MLLM to create \textit{personal information}. Since the cropped images contain only one person, the descriptions generated by the MLLM will only focus on the characteristics of this individual, which capture more fine-grained personalized details.

\paragraph{Holistic Information Extraction} As shown in Figure \ref{fig:framework}, although the personal features of the girl, such as ``single braid", can be captured through the previous step, without the holistic context, it would be impossible to extract the information ``the girl is looking at the man". Therefore, We utilize the existing descriptive capabilities of MLLMs to provide a \textit{holistic information} of the scene image. Specifically, we emphasize describing the main characters in the image, such as the holistic information of describing ``a man" and ``a girl" in Figure \ref{fig:framework}. This approach aims to offer more ``feature anchors" that facilitate the subsequent fusion of personal information and holistic information.

\paragraph{Dual-Level Information Fusion} After obtaining \textit{holistic information} that provides a contextual knowledge and \textit{personalized information} that captures individual characteristics, we attempt to link these two pieces of information. This is done by matching  the personal information with the descriptions of characters in the holistic information, which results in a fused description that describes how a specific individual interacts with the context (demonstrated in Appendix~\ref{tab:replace_person_prompt}). This dual-level fused information serves as the foundation for generating personalized conversations that are more detailed and contextually accurate.



\subsection{PVIT Dataset Generation}
With the visual concepts and the textual information associated with each individual extracted in the previous two phases , we can now construct the Personalized Visual Instruction Tuning (PVIT) dataset. The PVIT dataset primarily consists of three components: \textit{Personalized Multimodal Prefix}, \textit{Scene Image}, and \textit{Personalized QA}.

\subsubsection{Personalized Multimodal Prefix}
In our in-context learning formulation of the personalization task, each input individual is represented by a multimodal prefix, which is the combination of a personal image and a personalized introduction. The personal image can be drived from the previous stage, which can either be the headshot cropped from the original image, or a generated photo from Photomaker~\citep{li2023photomakercustomizingrealistichuman}. Th personalized introduction contains the essential knowledge that is related to the person, for which we consider the person's name. More advanced knowledge, such as character, hobbies and professions can be easily extended and will be considered in future work. We design strategies to diversify the coverage of personalized introduction.
\paragraph{Name Swapping} We introduce $<$name$>$ as a placeholder to be replaced with actual names during the construction of the training dataset.  Specifically, we collect a list of names (around 600 names) using ChatGPT. Then, we randomly select a name and swap it with the placeholder. It is important to note that this process can be repeated multiple times, augmenting the training data with diverse names. This approach not only enhances the model's generalization ability and robustness in adapting to new individuals, but also helps prevent overfitting by avoiding the association of a specific person with a fixed name during training. The effectiveness of this technique is verified in Section \ref{exp: repeat}.
\paragraph{Personal Pronoun} 
To better align with the way users refer to others in everyday conversations, we not only introduce individuals by their names but also handle situations involving personal pronouns (e.g., I, you, him). For example, if a question contains ``my dad," the response should adapt by using ``your dad." To handle such cases effectively, we introduce training data that includes examples of personal pronouns, ensuring the model can appropriately adjust its responses based on context.

\paragraph{Adversarial Introduction} To ensure the model truly learns to recognize individuals accurately, solely considering the cases where the questions are answerable is insufficient. The model must also learn to handle challenging or misleading scenarios. We found that even SOTA MLLMs often generate responses completely ignoring the input individuals provided in the prefix. For instance, even when the person of interest (e.g., a girl named Lisa) is not present in the image, model may still respond to questions about her and mistakenly identify other individuals as her. To address this issue, we introduce adversarial inputs designed to challenge the model's ability to correctly handle unanswerable questions. Specifically, we generate the following types of adversarial inputs:
\begin{itemize}
    \item \textbf{Adversarial name mapping}: When choosing the person of interest to construct the query (e.g., Lisa), we make sure that the person is not provided at all in the personalized multimodal prefixes. For this type of query, the model should respond with ``Sorry, I do not know who Lisa is", as the person was never introduced.
    \item \textbf{Adversarial image mapping}: When constructing the multimodal prefixes, we randomly select images of person excluding those that are in the scene image. The person of interest is one of the individual contained in the prefixes. In this case, the scene image does not contain this person of interest. The model should then respond with ``Sorry, I cannot see Lisa in the image", demonstrating its ability to correctly identify missing individuals.
\end{itemize}

\subsubsection{Scene Image}

In designing scene images containing specific individuals, our goal is to enable the model to accurately recognize the target individuals provided in the prefix. We design two types of scene images for this purpose. The first type is the original, complete image, with the full context containing additional elements besides the person. This allows subsequent Q\&A to be more comprehensive, focusing on both the specific features of individuals and their broader interactions with the environment. Additionally, to enhance the model's ability to recognize a specific person in an image containing multiple individuals, we also concatenate cropped images of individuals to create composite images, which elevates the MLLM's capability in more challenging scenarios.

\subsubsection{Personalized QA}



The dual-level information extracted from the previous stage transform visual information into texts, enabling us to utilize the LLM's advanced reasoning capabilities to create personalized conversations. 
Provided with such information, we meticulously design prompts and in-context examples to generate data for the following tasks using the LLM:
\begin{itemize}
    \item \textbf{Personalized Description}: We create conversations where the user queries for the description of specific individuals, rather than  the  entire  image.
    \item \textbf{Personalized Free-form QA}: We create free-form QA pairs that queries for information or characteristics related to specific individuals, such as appearance and behavior, which aims at enabling the MLLM to conduct personalized multi-round conversation.
    \item \textbf{Personalized Multi-choice QA} We create personalized QA pairs in multi-choice format, where the choices include the ground truth answer, as well as some confounding alternatives.
\end{itemize}
 Compared with general purpose image captioning and VQA, the personalized conversation presents novel challenges for current MLLMs, since they not only need to recognize the individuals of interest in the scene image, but also  properly incorporate them into the generated response.
 


In total, we create 3M training instances of personalized conversation, which we term \textbf{\Data}. The curated dataset encompasses diverse types of data and difficulty levels. We associate the detailed information and statistics of \Data in Section~\ref{sec:data_construction} of the Appendix..

\section{Evaluation Using \Bench}
Although numerous benchmarks have been proposed to assess the effectiveness of MLLMs, none have been specifically designed to evaluate their personalization capabilities. To address this gap, we present a high-quality benchmark manually checked by human, \textbf{\Bench}, aimed at thoroughly evaluating the personalization potential of MLLMs. We design both multiple choice (MC) questions and personalized image description queries for evaluation. In this section, we provide detailed overview of the problem types and evaluation metrics  of \Bench. Detailed statistics and curation process of the benchmark are presented in Section~\ref{sec:bench_stats}.

\subsection{Multiple-choice (MC) Questions}
We design positive (answerable) and adversarial (unanswerable) questions to examine the MLLM's ability to correctly associate the target individual with the corresponding person in the scene image.

\textbf{Answerable Questions} include the following types: 1) \textit{Crop}: the input individual is represented by its original face cropped from the image; 2) \textit{Aug-In}: using Photomaker~\citep{li2023photomakercustomizingrealistichuman}, we generate an augmented photo of individuals based on the original cropped face; 3) \textit{Aug-Sc-2} and \textit{Aug-Sc-3}: we concat two or three different cropped images of individuals into a single image to replace the original scene image, increasing the difficulty of accurately recognizing the individual.

\textbf{Unanswerable Questions}, on the other hand, include: 1) \textit{Adv-name}: the question pertains to a person who is not included in the list of input individuals, meaning the MLLM lacks knowledge of this person; 2) \textit{Adv-image}: the individual mentioned in the question does not appear in the scene image, meaning the MLLM cannot visually identify this person.

\paragraph{Evaluation Metrics} For MC questions, we adopt accuracy for the evaluation metric. To gain a deeper understanding of the MLLM's capabilities and limitations, we separately evaluate each of the three types in MC questions. In addition, to further study the MLLM's ability to differentiate different individuals, we also separately evaluate the accuracy of images containing different numbers of people.
\subsection{Descriptive Questions}
For this type of questions, we query for the descriptions of specific individuals, rather than general descriptions. We also design both positive and adversarial description questions. Specifically, the positive questions involves different number of people in the scene images. As the number of people grows in the scene image, it becomes more challenging for the MLLM to correctly recognize the person of interest and produce accurate descriptions.

\paragraph{Evaluation Metrics} For descriptions, we perform evaluations using different strategies for answerable and unanswerable questions. For answerable ones, we adopt LongClip~\citep{zhang2024longclipunlockinglongtextcapability} to evaluate the similarity between the image of the target individual and the MLLM-generated descriptions. For unanswerable ones, we calculate the percentage that the MLLM rejects to respond.


\section{Experiments}
In this section, we demonstrate the effectiveness of our proposed \ModelNameAbbre on the constructed \Bench. We first showcase the results of the \ModelNameAbbre-tuned LLaVA~\citep{liu2023llava}, which demonstrates significantly higher performances than the SOTA MLLMs that support multi-image inputs. Then, we conduct ablation study on each of the components of our constructed data to demonstrate their contributions towards the final performance. We provide the detailed training configurations in Section~\ref{sec:training_detail} of the Appendix.

\begin{table}[t!]
\caption{MC questions on \Bench. P-LLaVA trained with \ModelNameAbbre significantly outperforms other MLLMs across various question types. Remarkably, P-LLaVA demonstrates strong performances on challenging answerable tasks and unanswerable queries, where the other MLLMs drastically fail.
}
\label{tab:main_results_mc}
    \centering
    \resizebox{\textwidth}{!}
{
\begin{tabular}{c!{\vrule width 0.5pt}ccccc!{\vrule width 0.5pt}ccc}
\toprule
 & \multicolumn{5}{c}{Answerable}& \multicolumn{3}{c}{Unanswerable}\\
MLLM & Crop & Aug-In & Aug-Sc-2 & Aug-Sc-3 & Avg & Adv-Img & Adv-Name & Avg \\
\midrule
Qwen-VL-7B~\citep{bai2023qwenvl} & 66.58& 72.92 & 60.87 & 50.57 & 62.74 & 1.11 & 20.62 & 10.87 \\
VILA1.5-7B~\citep{lin2024vila} & 57.54& 48.39 & 55.91 & 38.29 & 50.03 & 36.36 & 3.23 & 19.79\\
LLaVA-OneVision-7B~\citep{li2024llavaonevisioneasyvisualtask} & 84.82 & 82.17& 77.53 & 78.31 & 80.71 & 24.12 & 1.07 & 12.60\\
InternVL-Chat-V1.5-26B~\citep{chen2024internvl} & 62.99 & 56.84& 57.14 & 51.02& 57.00& 53.33 & 9.64 & 31.49\\
Deepseek-VL-7b-chat~\citep{lu2024deepseek} & 89.72 & 24.79 & 71.43 & 66.56 & 63.13 & 4.12 & 0.00 & 2.06\\
mPLUG-OWl2~\citep{ye2024mplug} & 72.29 & 70.21 & 61.34 & 50.63 & 63.62 & 0.00 & 5.00 & 2.50\\
\midrule
P-LLaVA (Ours) & \cc \textbf{95.42} & \cc \textbf{97.23} & \cc \textbf{96.48} & \cc \textbf{97.62} & \cc \textbf{96.69} & \cc \textbf{99.43} & \cc \textbf{100.00} & \cc \textbf{99.72}\\
\bottomrule
\end{tabular}
}

\end{table}

\begin{table}
\caption{Evaluation of personalized description capability on \Bench. The P-LLaVA tuned with our \ModelNameAbbre is able to accurately recognize and describe the person of interest, even for challenging cases where multiple individuals are contained in the image.
}
\label{tab:main_results_des}
    \centering
    \resizebox{\textwidth}{!}
{
\begin{tabular}{c!{\vrule width 0.5pt}ccccc!{\vrule width 0.5pt}ccc!{\vrule width 0.5pt}}
\toprule
 & \multicolumn{5}{c}{Answerable}& \multicolumn{3}{c}{Unanswerable}\\
MLLM & Cnt=1 & Cnt=2 & Cnt=3 & Cnt$>=$4 & Avg & Adv-Img & Adv-Name & Avg \\
\midrule
Qwen-VL-7B~\citep{bai2023qwenvl} & 76.20& 72.38 & 69.59 & 63.04 & 70.30 & 0.00 & 15.00 & 7.50 \\
VILA1.5-7B~\citep{lin2024vila} & 74.90& 74.58 & 67.29 & 70.02 & 71.70 & 10.00 & 20.00 & 15.00\\
LLaVA-OneVision-7B~\citep{li2024llavaonevisioneasyvisualtask} & 74.54 & 71.08& 76.13 & 63.54 & 71.32 & 10.00 & 15.00 & 12.50\\
InternVL-Chat-V1.5-26B~\citep{chen2024internvl} & 82.64 & 75.94 & 67.12 & 71.28 & 74.25 & 15.00 & 20.00 & 17.50\\
Deepseek-VL-7b-chat~\citep{lu2024deepseek} & 82.15 & 77.86 & 76.06 & 76.44 & 78.13 & 0.00 & 5.00 & 2.50\\
mPLUG-OWl2~\citep{ye2024mplug} & 80.38 & 79.11 & 77.77 & 72.68 & 77.49 & 0.00 & 5.00 & 2.50\\

\midrule
P-LLaVA (Ours) & \cc \textbf{85.24} & \cc \textbf{83.10} & \cc \textbf{83.14} & \cc \textbf{78.71} & \cc \textbf{82.55} & \cc \textbf{100.00} & \cc \textbf{100.00} & \cc \textbf{100.00}\\
\bottomrule
\end{tabular}
}

\end{table}

\subsection{Main Results on \Bench}
We compare the personalization capability of P-LLaVA trained with our \ModelNameAbbre with other SOTA MLLMs on our constructed \Bench. We conduct evaluations using both the MC questions and personalized descriptions in Table~\ref{tab:main_results_mc} and  Table~\ref{tab:main_results_des}, respectively. We observe the following phenomena for current SOTA MLLMs: 



\textbf{1)} The performances of SOTA MLLMs are significantly lower with more complex inputs (i.e., Aug-In, Aug-Sc-2 and Aug-Sc-3 for MC questions, and scene images that contain more people for description questions), which indicates their limited capability and robustness in recognizing input individuals for the scene images. 

\textbf{2)} All the MLLMs drastically fail for unanswerable questions. They still tend to answer the questions that are not answerable by mistakenly treating other people in the image as the person of interest. This is because the MLLMs have never been trained to reject replying to such unanswerable questions.

\textbf{3)} After fine-tuning LLaVA~\citep{liu2023llava} with our \ModelNameAbbre, we observe significant performance boosts for all question types in \Bench. Specifically, we observe performance enhancement for both positive and unanswerable questions. Notably, the performance on more complex scene images is boosted even more significantly. The results verifies the effectiveness of our propose tuning strategy in improving the model's personalization capability.

\subsection{Ablation Study}
\begin{table}[h]
\begin{minipage}[c]{0.62\textwidth}
\caption{\small{The performances on MC questions with scene images containing different numbers of people.}}
\label{tab:count_ablation}
    \centering
    \resizebox{1\textwidth}{!}
{
\begin{tabular}{c!{\vrule width 0.5pt}ccccc!{\vrule width 0.5pt}ccc!{\vrule width 0.5pt}}
\toprule
MLLM & Cnt=1 & Cnt=2 & Cnt=3 & Cnt$>=$4 \\
\midrule
Qwen-VL-7B~\citep{bai2023qwenvl} & 74.67& 58.71 & 57.38 & 37.32 \\
VILA1.5-7B~\citep{lin2024vila} & 64.31& 47.02 & 44.61 & 38.13\\
LLaVA-OneVision-7B~\citep{li2024llavaonevisioneasyvisualtask} & 88.72 & 83.23& 80.37 & 76.31\\
InternVL-Chat-V1.5-26B~\citep{chen2024internvl} & 56.21 & 52.28 & 43.57 & 38.31\\
Deepseek-VL-7b-chat~\citep{lu2024deepseek} & 84.03 & 68.71 & 76.52 & 74.90\\
mPLUG-OWl2~\citep{ye2024mplug} & 78.90 & 63.23 & 77.43 & 60.51\\

\midrule
P-LLaVA (Ours) & \cc \textbf{98.71} & \cc \textbf{95.03} & \cc \textbf{94.90} & \cc \textbf{95.32}\\
\bottomrule
\end{tabular}
}
\end{minipage}
\hspace{0.5cm}
\begin{minipage}[c]{0.36 \textwidth}
\centering
\caption{\small{Face augmentation boosts the MLLM's capabilities to recoginize individuals, while adversarial samples are critical for enabling  MLLM to reject answering the unanswerable questions.
}
}
\label{tab:data_ablation}

\resizebox{\textwidth}{!}{
\begin{tabular}{c!{\vrule width 0.5pt}cccc}
\toprule
Data & Crop & Augment & Unanswerable\\
\midrule
{Full} & \cc \textbf{96.43} & \cc \textbf{95.32} & \cc \textbf{99.78}  \\
{wo Aug}& 94.93 & 88.43 & 98.73 \\
{wo Adv}& 95.92 & 92.48 & 1.12 \\
\bottomrule
\end{tabular}}
\end{minipage}

\end{table}



\subsubsection{Performance for vairous Number of person} 
In Table~\ref{tab:count_ablation}, we demonstrate the MLLMs' personalization performances on MC questions with scene images containing different numbers of people. The SOTA MLLMs showcase deteriorated performances on scene images containing more people due to the challenge of accurately recognizing the specific individual of interest. On the other hand, our trained P-LLaVA remains highly accurate on such challenging cases,  which verifies its capability to accurately recognize person of interest.
 
\subsubsection{data ablation} We examine the effectiveness of each data component in our curated dataset in Table~\ref{tab:data_ablation}. We observe the following: 1) Face augmentation using PhotoMaker increases the diversity of  input individuals, which effectively boosts the MLLM's capabilities to recoginize individuals;  2) Adversarial samples are critical for enabling  MLLM to reject answering queries that are unanswerable. Without adversarial samples, the accuracy for rejecting answering unanswerable questions rapidly drops to near zero.
\subsubsection{Impact of Data Scale and Name Repetition}
\label{exp: repeat}
\begin{wrapfigure}[10]{r}{0.4\textwidth}
\begin{minipage}{.4\textwidth}
\vspace{-6mm}
\includegraphics[width=1\textwidth]{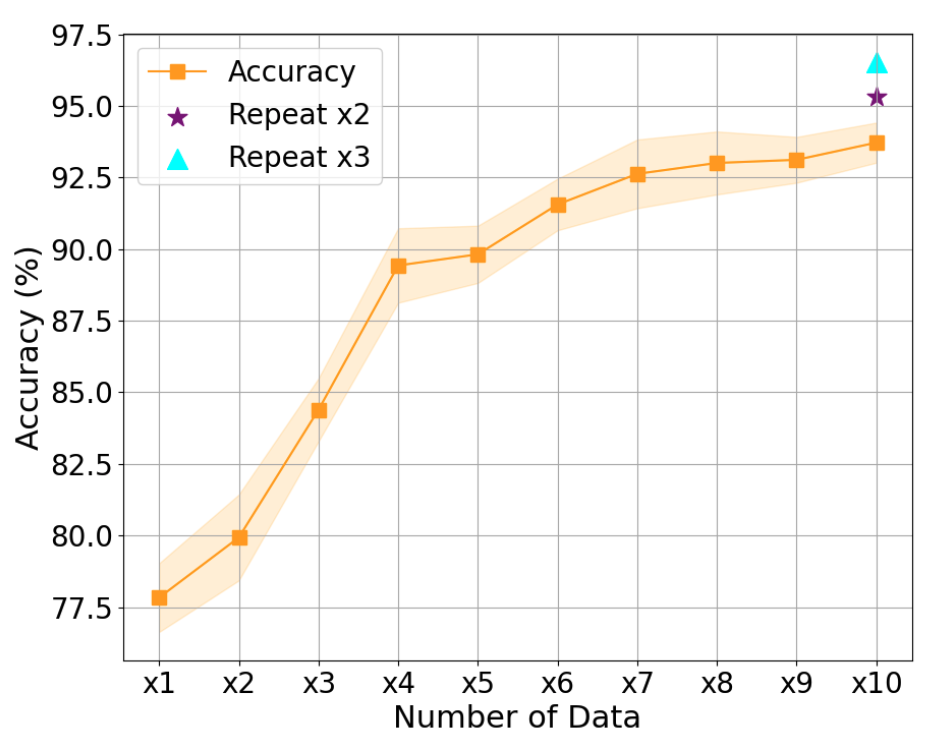} 
\end{minipage}
\end{wrapfigure}
In the figure on the right, we illustrate the evaluation accuracy after training with various amounts of data. Specifically, the horizontal axis indicates the number of data units, and each unit contains 8000 samples. We observe clear performance boost when scaling up the training dataset. Furthermore, we find that even with the same data templates, by repeatedly constructing data using different names, the performance of MLLM is able to be further enhanced, which verifies that the diversity of names used during training makes the personalization capability more robust and generalizable to new individuals.
\section{Conclusion}
In this work, we introduce \ModelName (\ModelNameAbbre), a novel formulation a training framework that enables personalized conversations targeting specific individuals. To achieve this, We first develop an automatic pipeline that generates training data with individuals in diverse visual and conversational contexts. Then, we adopt the data to finetune the MLLM, which significantly improves MLLM's personalized dialogue capabilities, as demonstrated through the \Bench benchmark. We hope that our work will promote the advancement in personalized applications, such as visual assistants and domestic robots, enhancing user-centric multimodal interactions.

\bibliography{arxiv}

\begin{thebibliography}{55}
\providecommand{\natexlab}[1]{#1}
\providecommand{\url}[1]{\texttt{#1}}
\expandafter\ifx\csname urlstyle\endcsname\relax
  \providecommand{\doi}[1]{doi: #1}\else
  \providecommand{\doi}{doi: \begingroup \urlstyle{rm}\Url}\fi

\bibitem[Alaluf et~al.(2023)Alaluf, Richardson, Metzer, and Cohen-Or]{alaluf2023neural}
Yuval Alaluf, Elad Richardson, Gal Metzer, and Daniel Cohen-Or.
\newblock A neural space-time representation for text-to-image personalization, 2023.

\bibitem[Alaluf et~al.(2024)Alaluf, Richardson, Tulyakov, Aberman, and Cohen-Or]{alaluf2024myvlmpersonalizingvlmsuserspecific}
Yuval Alaluf, Elad Richardson, Sergey Tulyakov, Kfir Aberman, and Daniel Cohen-Or.
\newblock Myvlm: Personalizing vlms for user-specific queries, 2024.
\newblock URL \url{https://arxiv.org/abs/2403.14599}.

\bibitem[Arar et~al.(2023)Arar, Gal, Atzmon, Chechik, Cohen-Or, Shamir, and Bermano]{arar2023domainagnostic}
Moab Arar, Rinon Gal, Yuval Atzmon, Gal Chechik, Daniel Cohen-Or, Ariel Shamir, and Amit~H. Bermano.
\newblock Domain-agnostic tuning-encoder for fast personalization of text-to-image models, 2023.

\bibitem[Bai et~al.(2023)Bai, Bai, Yang, Wang, Tan, Wang, Lin, Zhou, and Zhou]{bai2023qwenvl}
Jinze Bai, Shuai Bai, Shusheng Yang, Shijie Wang, Sinan Tan, Peng Wang, Junyang Lin, Chang Zhou, and Jingren Zhou.
\newblock Qwen-vl: A versatile vision-language model for understanding, localization, text reading, and beyond, 2023.

\bibitem[Bai et~al.(2022)Bai, Jones, Ndousse, Askell, Chen, DasSarma, Drain, Fort, Ganguli, Henighan, et~al.]{bai2022training}
Yuntao Bai, Andy Jones, Kamal Ndousse, Amanda Askell, Anna Chen, Nova DasSarma, Dawn Drain, Stanislav Fort, Deep Ganguli, Tom Henighan, et~al.
\newblock Training a helpful and harmless assistant with reinforcement learning from human feedback.
\newblock \emph{arXiv preprint arXiv:2204.05862}, 2022.

\bibitem[Brown et~al.(2020)Brown, Mann, Ryder, Subbiah, Kaplan, Dhariwal, Neelakantan, Shyam, Sastry, Askell, et~al.]{brown2020language}
Tom Brown, Benjamin Mann, Nick Ryder, Melanie Subbiah, Jared~D Kaplan, Prafulla Dhariwal, Arvind Neelakantan, Pranav Shyam, Girish Sastry, Amanda Askell, et~al.
\newblock Language models are few-shot learners.
\newblock \emph{Advances in neural information processing systems}, 33:\penalty0 1877--1901, 2020.

\bibitem[Chen et~al.(2024)Chen, Wu, Wang, Su, Chen, Xing, Zhong, Zhang, Zhu, Lu, et~al.]{chen2024internvl}
Zhe Chen, Jiannan Wu, Wenhai Wang, Weijie Su, Guo Chen, Sen Xing, Muyan Zhong, Qinglong Zhang, Xizhou Zhu, Lewei Lu, et~al.
\newblock Internvl: Scaling up vision foundation models and aligning for generic visual-linguistic tasks.
\newblock In \emph{Proceedings of the IEEE/CVF Conference on Computer Vision and Pattern Recognition}, pp.\  24185--24198, 2024.

\bibitem[Chiang et~al.(2023)Chiang, Li, Lin, Sheng, Wu, Zhang, Zheng, Zhuang, Zhuang, Gonzalez, Stoica, and Xing]{vicuna2023}
Wei-Lin Chiang, Zhuohan Li, Zi~Lin, Ying Sheng, Zhanghao Wu, Hao Zhang, Lianmin Zheng, Siyuan Zhuang, Yonghao Zhuang, Joseph~E. Gonzalez, Ion Stoica, and Eric~P. Xing.
\newblock Vicuna: An open-source chatbot impressing gpt-4 with 90\%* chatgpt quality, March 2023.
\newblock URL \url{https://lmsys.org/blog/2023-03-30-vicuna/}.

\bibitem[Chowdhery et~al.(2022)Chowdhery, Narang, Devlin, Bosma, Mishra, Roberts, Barham, Chung, Sutton, Gehrmann, et~al.]{chowdhery2022palm}
Aakanksha Chowdhery, Sharan Narang, Jacob Devlin, Maarten Bosma, Gaurav Mishra, Adam Roberts, Paul Barham, Hyung~Won Chung, Charles Sutton, Sebastian Gehrmann, et~al.
\newblock Palm: Scaling language modeling with pathways.
\newblock \emph{arXiv preprint arXiv:2204.02311}, 2022.

\bibitem[Chunseong~Park et~al.(2017)Chunseong~Park, Kim, and Kim]{chunseong2017attend}
Cesc Chunseong~Park, Byeongchang Kim, and Gunhee Kim.
\newblock Attend to you: Personalized image captioning with context sequence memory networks.
\newblock In \emph{Proceedings of the IEEE conference on computer vision and pattern recognition}, pp.\  895--903, 2017.

\bibitem[Dai et~al.(2023)Dai, Li, Li, Tiong, Zhao, Wang, Li, Fung, and Hoi]{dai2023instructblip}
Wenliang Dai, Junnan Li, Dongxu Li, Anthony Meng~Huat Tiong, Junqi Zhao, Weisheng Wang, Boyang Li, Pascale Fung, and Steven Hoi.
\newblock Instructblip: Towards general-purpose vision-language models with instruction tuning, 2023.

\bibitem[Ding et~al.(2023)Ding, Han, Xu, Zhang, and Li]{ding2023hilm}
Xinpeng Ding, Jianhua Han, Hang Xu, Wei Zhang, and Xiaomeng Li.
\newblock Hilm-d: Towards high-resolution understanding in multimodal large language models for autonomous driving.
\newblock \emph{arXiv preprint arXiv:2309.05186}, 2023.

\bibitem[Gal et~al.(2023)Gal, Alaluf, Atzmon, Patashnik, Bermano, Chechik, and Cohen-or]{gal2023image}
Rinon Gal, Yuval Alaluf, Yuval Atzmon, Or~Patashnik, Amit~Haim Bermano, Gal Chechik, and Daniel Cohen-or.
\newblock An image is worth one word: Personalizing text-to-image generation using textual inversion.
\newblock In \emph{The Eleventh International Conference on Learning Representations}, 2023.
\newblock URL \url{https://openreview.net/forum?id=NAQvF08TcyG}.

\bibitem[Gao et~al.(2023{\natexlab{a}})Gao, Pi, Lin, Xu, Ye, Wu, Zhang, Liang, Li, and Kong]{gao2023selfguided}
Jiahui Gao, Renjie Pi, Yong Lin, Hang Xu, Jiacheng Ye, Zhiyong Wu, Weizhong Zhang, Xiaodan Liang, Zhenguo Li, and Lingpeng Kong.
\newblock Self-guided noise-free data generation for efficient zero-shot learning, 2023{\natexlab{a}}.

\bibitem[Gao et~al.(2023{\natexlab{b}})Gao, Pi, Zhang, Ye, Zhong, Wang, Hong, Han, Xu, Li, and Kong]{gao2023gllava}
Jiahui Gao, Renjie Pi, Jipeng Zhang, Jiacheng Ye, Wanjun Zhong, Yufei Wang, Lanqing Hong, Jianhua Han, Hang Xu, Zhenguo Li, and Lingpeng Kong.
\newblock G-llava: Solving geometric problem with multi-modal large language model, 2023{\natexlab{b}}.

\bibitem[Gao et~al.(2023{\natexlab{c}})Gao, Han, Zhang, Lin, Geng, Zhou, Zhang, Lu, He, Yue, Li, and Qiao]{gao2023llamaadapter}
Peng Gao, Jiaming Han, Renrui Zhang, Ziyi Lin, Shijie Geng, Aojun Zhou, Wei Zhang, Pan Lu, Conghui He, Xiangyu Yue, Hongsheng Li, and Yu~Qiao.
\newblock Llama-adapter v2: Parameter-efficient visual instruction model, 2023{\natexlab{c}}.

\bibitem[Geitgey(2016)]{geitgey2016machine}
Adam Geitgey.
\newblock Machine learning is fun! part 4: Modern face recognition with deep learning.
\newblock \emph{Medium. Medium Corporation}, 24:\penalty0 2016, 2016.

\bibitem[Geng \& Liu(2023)Geng and Liu]{openlm2023openllama}
Xinyang Geng and Hao Liu.
\newblock Openllama: An open reproduction of llama, May 2023.
\newblock URL \url{https://github.com/openlm-research/open_llama}.

\bibitem[Hoffmann et~al.(2022)Hoffmann, Borgeaud, Mensch, Buchatskaya, Cai, Rutherford, Casas, Hendricks, Welbl, Clark, et~al.]{hoffmann2022training}
Jordan Hoffmann, Sebastian Borgeaud, Arthur Mensch, Elena Buchatskaya, Trevor Cai, Eliza Rutherford, Diego de~Las Casas, Lisa~Anne Hendricks, Johannes Welbl, Aidan Clark, et~al.
\newblock Training compute-optimal large language models.
\newblock \emph{arXiv preprint arXiv:2203.15556}, 2022.

\bibitem[Li et~al.(2024)Li, Zhang, Guo, Zhang, Li, Zhang, Zhang, Li, Liu, and Li]{li2024llavaonevisioneasyvisualtask}
Bo~Li, Yuanhan Zhang, Dong Guo, Renrui Zhang, Feng Li, Hao Zhang, Kaichen Zhang, Yanwei Li, Ziwei Liu, and Chunyuan Li.
\newblock Llava-onevision: Easy visual task transfer, 2024.
\newblock URL \url{https://arxiv.org/abs/2408.03326}.

\bibitem[Li et~al.(2023{\natexlab{a}})Li, Wong, Zhang, Usuyama, Liu, Yang, Naumann, Poon, and Gao]{li2023llavamed}
Chunyuan Li, Cliff Wong, Sheng Zhang, Naoto Usuyama, Haotian Liu, Jianwei Yang, Tristan Naumann, Hoifung Poon, and Jianfeng Gao.
\newblock Llava-med: Training a large language-and-vision assistant for biomedicine in one day, 2023{\natexlab{a}}.

\bibitem[Li et~al.(2023{\natexlab{b}})Li, Li, Savarese, and Hoi]{li2023blip2}
Junnan Li, Dongxu Li, Silvio Savarese, and Steven Hoi.
\newblock Blip-2: Bootstrapping language-image pre-training with frozen image encoders and large language models, 2023{\natexlab{b}}.

\bibitem[Li et~al.(2023{\natexlab{c}})Li, Cao, Wang, Qi, Cheng, and Shan]{li2023photomakercustomizingrealistichuman}
Zhen Li, Mingdeng Cao, Xintao Wang, Zhongang Qi, Ming-Ming Cheng, and Ying Shan.
\newblock Photomaker: Customizing realistic human photos via stacked id embedding, 2023{\natexlab{c}}.
\newblock URL \url{https://arxiv.org/abs/2312.04461}.

\bibitem[Lin et~al.(2024)Lin, Yin, Ping, Molchanov, Shoeybi, and Han]{lin2024vila}
Ji~Lin, Hongxu Yin, Wei Ping, Pavlo Molchanov, Mohammad Shoeybi, and Song Han.
\newblock Vila: On pre-training for visual language models.
\newblock In \emph{Proceedings of the IEEE/CVF Conference on Computer Vision and Pattern Recognition}, pp.\  26689--26699, 2024.

\bibitem[Liu et~al.(2023{\natexlab{a}})Liu, Li, Wu, and Lee]{liu2023llava}
Haotian Liu, Chunyuan Li, Qingyang Wu, and Yong~Jae Lee.
\newblock Visual instruction tuning, 2023{\natexlab{a}}.

\bibitem[Liu et~al.(2023{\natexlab{b}})Liu, Zeng, Ren, Li, Zhang, Yang, Li, Yang, Su, Zhu, and Zhang]{liu2023grounding}
Shilong Liu, Zhaoyang Zeng, Tianhe Ren, Feng Li, Hao Zhang, Jie Yang, Chunyuan Li, Jianwei Yang, Hang Su, Jun Zhu, and Lei Zhang.
\newblock Grounding dino: Marrying dino with grounded pre-training for open-set object detection, 2023{\natexlab{b}}.

\bibitem[Lu et~al.(2024)Lu, Liu, Zhang, Wang, Dong, Liu, Sun, Ren, Li, Sun, et~al.]{lu2024deepseek}
Haoyu Lu, Wen Liu, Bo~Zhang, Bingxuan Wang, Kai Dong, Bo~Liu, Jingxiang Sun, Tongzheng Ren, Zhuoshu Li, Yaofeng Sun, et~al.
\newblock Deepseek-vl: towards real-world vision-language understanding.
\newblock \emph{arXiv preprint arXiv:2403.05525}, 2024.

\bibitem[Meng et~al.(2022)Meng, Huang, Zhang, and Han]{meng2022generating}
Yu~Meng, Jiaxin Huang, Yu~Zhang, and Jiawei Han.
\newblock Generating training data with language models: Towards zero-shot language understanding.
\newblock \emph{arXiv preprint arXiv:2202.04538}, 2022.

\bibitem[Meta(2024)]{dubey2024llama3herdmodels}
Meta.
\newblock The llama 3 herd of models, 2024.
\newblock URL \url{https://arxiv.org/abs/2407.21783}.

\bibitem[Nguyen et~al.(2024)Nguyen, Liu, Li, Cai, Ojha, and Lee]{nguyen2024yollavapersonalizedlanguagevision}
Thao Nguyen, Haotian Liu, Yuheng Li, Mu~Cai, Utkarsh Ojha, and Yong~Jae Lee.
\newblock Yo'llava: Your personalized language and vision assistant, 2024.
\newblock URL \url{https://arxiv.org/abs/2406.09400}.

\bibitem[OpenAI(2023)]{openai2023gpt4}
OpenAI.
\newblock Gpt-4 technical report, 2023.

\bibitem[Ouyang et~al.(2022)Ouyang, Wu, Jiang, Almeida, Wainwright, Mishkin, Zhang, Agarwal, Slama, Ray, et~al.]{ouyang2022training}
Long Ouyang, Jeffrey Wu, Xu~Jiang, Diogo Almeida, Carroll Wainwright, Pamela Mishkin, Chong Zhang, Sandhini Agarwal, Katarina Slama, Alex Ray, et~al.
\newblock Training language models to follow instructions with human feedback.
\newblock \emph{Advances in Neural Information Processing Systems}, 35:\penalty0 27730--27744, 2022.

\bibitem[Park et~al.(2018)Park, Kim, and Kim]{park2018towards}
Cesc~Chunseong Park, Byeongchang Kim, and Gunhee Kim.
\newblock Towards personalized image captioning via multimodal memory networks.
\newblock \emph{IEEE transactions on pattern analysis and machine intelligence}, 41\penalty0 (4):\penalty0 999--1012, 2018.

\bibitem[Pi et~al.(2023{\natexlab{a}})Pi, Gao, Diao, Pan, Dong, Zhang, Yao, Han, Xu, Kong, and Zhang]{pi2023detgpt}
Renjie Pi, Jiahui Gao, Shizhe Diao, Rui Pan, Hanze Dong, Jipeng Zhang, Lewei Yao, Jianhua Han, Hang Xu, Lingpeng Kong, and Tong Zhang.
\newblock Detgpt: Detect what you need via reasoning, 2023{\natexlab{a}}.

\bibitem[Pi et~al.(2023{\natexlab{b}})Pi, Yao, Gao, Zhang, and Zhang]{pi2023perceptiongpt}
Renjie Pi, Lewei Yao, Jiahui Gao, Jipeng Zhang, and Tong Zhang.
\newblock Perceptiongpt: Effectively fusing visual perception into llm, 2023{\natexlab{b}}.

\bibitem[Pi et~al.(2024{\natexlab{a}})Pi, Han, Xie, Pan, Lian, Dong, Zhang, and Zhang]{pi2024mllmprotector}
Renjie Pi, Tianyang Han, Yueqi Xie, Rui Pan, Qing Lian, Hanze Dong, Jipeng Zhang, and Tong Zhang.
\newblock Mllm-protector: Ensuring mllm's safety without hurting performance, 2024{\natexlab{a}}.

\bibitem[Pi et~al.(2024{\natexlab{b}})Pi, Han, Xiong, Zhang, Liu, Pan, and Zhang]{pi2024strengthening}
Renjie Pi, Tianyang Han, Wei Xiong, Jipeng Zhang, Runtao Liu, Rui Pan, and Tong Zhang.
\newblock Strengthening multimodal large language model with bootstrapped preference optimization.
\newblock \emph{arXiv preprint arXiv:2403.08730}, 2024{\natexlab{b}}.

\bibitem[Pi et~al.(2024{\natexlab{c}})Pi, Zhang, Zhang, Pan, Chen, and Zhang]{pi2024image}
Renjie Pi, Jianshu Zhang, Jipeng Zhang, Rui Pan, Zhekai Chen, and Tong Zhang.
\newblock Image textualization: An automatic framework for creating accurate and detailed image descriptions.
\newblock \emph{arXiv preprint arXiv:2406.07502}, 2024{\natexlab{c}}.

\bibitem[Ruiz et~al.(2022)Ruiz, Li, Jampani, Pritch, Rubinstein, and Aberman]{ruiz2022dreambooth}
Nataniel Ruiz, Yuanzhen Li, Varun Jampani, Yael Pritch, Michael Rubinstein, and Kfir Aberman.
\newblock Dreambooth: Fine tuning text-to-image diffusion models for subject-driven generation.
\newblock 2022.

\bibitem[Scao et~al.(2022)Scao, Fan, Akiki, Pavlick, Ili{c}, Hesslow, Castagn{e}, Luccioni, Yvon, Gall{e}, et~al.]{scao2022bloom}
Teven~Le Scao, Angela Fan, Christopher Akiki, Ellie Pavlick, Suzana Ili{c}, Daniel Hesslow, Roman Castagn{e}, Alexandra~Sasha Luccioni, Fran{\c{c}}ois Yvon, Matthias Gall{e}, et~al.
\newblock Bloom: A 176b-parameter open-access multilingual language model.
\newblock \emph{arXiv preprint arXiv:2211.05100}, 2022.

\bibitem[Shuster et~al.(2019)Shuster, Humeau, Hu, Bordes, and Weston]{shuster2019engaging}
Kurt Shuster, Samuel Humeau, Hexiang Hu, Antoine Bordes, and Jason Weston.
\newblock Engaging image captioning via personality.
\newblock In \emph{Proceedings of the IEEE/CVF Conference on Computer Vision and Pattern Recognition}, pp.\  12516--12526, 2019.

\bibitem[Smith et~al.(2022)Smith, Patwary, Norick, LeGresley, Rajbhandari, Casper, Liu, Prabhumoye, Zerveas, Korthikanti, et~al.]{smith2022using}
Shaden Smith, Mostofa Patwary, Brandon Norick, Patrick LeGresley, Samyam Rajbhandari, Jared Casper, Zhun Liu, Shrimai Prabhumoye, George Zerveas, Vijay Korthikanti, et~al.
\newblock Using deepspeed and megatron to train megatron-turing nlg 530b, a large-scale generative language model.
\newblock \emph{arXiv preprint arXiv:2201.11990}, 2022.

\bibitem[Su et~al.(2023)Su, Lan, Li, Xu, Wang, and Cai]{su2023pandagpt}
Yixuan Su, Tian Lan, Huayang Li, Jialu Xu, Yan Wang, and Deng Cai.
\newblock Pandagpt: One model to instruction-follow them all, 2023.

\bibitem[Taori et~al.(2023)Taori, Gulrajani, Zhang, Dubois, Li, Guestrin, Liang, and Hashimoto]{alpaca}
Rohan Taori, Ishaan Gulrajani, Tianyi Zhang, Yann Dubois, Xuechen Li, Carlos Guestrin, Percy Liang, and Tatsunori~B. Hashimoto.
\newblock Stanford alpaca: An instruction-following llama model.
\newblock \url{https://github.com/tatsu-lab/stanford_alpaca}, 2023.

\bibitem[Touvron et~al.(2023)Touvron, Lavril, Izacard, Martinet, Lachaux, Lacroix, Rozi{\`e}re, Goyal, Hambro, Azhar, et~al.]{touvron2023llama}
Hugo Touvron, Thibaut Lavril, Gautier Izacard, Xavier Martinet, Marie-Anne Lachaux, Timoth{e}e Lacroix, Baptiste Rozi{\`e}re, Naman Goyal, Eric Hambro, Faisal Azhar, et~al.
\newblock Llama: Open and efficient foundation language models.
\newblock \emph{arXiv preprint arXiv:2302.13971}, 2023.

\bibitem[Voynov et~al.(2023)Voynov, Chu, Cohen-Or, and Aberman]{voynov2023p+}
Andrey Voynov, Qinghao Chu, Daniel Cohen-Or, and Kfir Aberman.
\newblock $ p+ $: Extended textual conditioning in text-to-image generation.
\newblock \emph{arXiv preprint arXiv:2303.09522}, 2023.

\bibitem[Wang et~al.(2024)Wang, Bai, Wang, Qin, and Chen]{wang2024instantid}
Qixun Wang, Xu~Bai, Haofan Wang, Zekui Qin, and Anthony Chen.
\newblock Instantid: Zero-shot identity-preserving generation in seconds.
\newblock \emph{arXiv preprint arXiv:2401.07519}, 2024.

\bibitem[Wang et~al.(2023)Wang, Wang, Chai, Zhou, and Wang]{wang2023user}
Xuan Wang, Guanhong Wang, Wenhao Chai, Jiayu Zhou, and Gaoang Wang.
\newblock User-aware prefix-tuning is a good learner for personalized image captioning.
\newblock In \emph{Chinese Conference on Pattern Recognition and Computer Vision (PRCV)}, pp.\  384--395. Springer, 2023.

\bibitem[Wei et~al.(2023)Wei, Wang, Schuurmans, Bosma, Ichter, Xia, Chi, Le, and Zhou]{wei2023chainofthoughtpromptingelicitsreasoning}
Jason Wei, Xuezhi Wang, Dale Schuurmans, Maarten Bosma, Brian Ichter, Fei Xia, Ed~Chi, Quoc Le, and Denny Zhou.
\newblock Chain-of-thought prompting elicits reasoning in large language models, 2023.
\newblock URL \url{https://arxiv.org/abs/2201.11903}.

\bibitem[Ye et~al.(2023)Ye, Zhang, Liu, Han, and Yang]{ye2023ip-adapter}
Hu~Ye, Jun Zhang, Sibo Liu, Xiao Han, and Wei Yang.
\newblock Ip-adapter: Text compatible image prompt adapter for text-to-image diffusion models.
\newblock 2023.

\bibitem[Ye et~al.(2022)Ye, Gao, Li, Xu, Feng, Wu, Yu, and Kong]{ye2022zerogen}
Jiacheng Ye, Jiahui Gao, Qintong Li, Hang Xu, Jiangtao Feng, Zhiyong Wu, Tao Yu, and Lingpeng Kong.
\newblock Zerogen: Efficient zero-shot learning via dataset generation.
\newblock In \emph{Empirical Methods in Natural Language Processing}, 2022.

\bibitem[Ye et~al.(2024)Ye, Xu, Ye, Yan, Hu, Liu, Qian, Zhang, and Huang]{ye2024mplug}
Qinghao Ye, Haiyang Xu, Jiabo Ye, Ming Yan, Anwen Hu, Haowei Liu, Qi~Qian, Ji~Zhang, and Fei Huang.
\newblock mplug-owl2: Revolutionizing multi-modal large language model with modality collaboration.
\newblock In \emph{Proceedings of the IEEE/CVF Conference on Computer Vision and Pattern Recognition}, pp.\  13040--13051, 2024.

\bibitem[Zeng et~al.(2019)Zeng, Abuduweili, Li, and Yang]{zeng2019automatic}
Wenhuan Zeng, Abulikemu Abuduweili, Lei Li, and Pengcheng Yang.
\newblock Automatic generation of personalized comment based on user profile.
\newblock \emph{arXiv preprint arXiv:1907.10371}, 2019.

\bibitem[Zhang et~al.(2024)Zhang, Zhang, Dong, Zang, and Wang]{zhang2024longclipunlockinglongtextcapability}
Beichen Zhang, Pan Zhang, Xiaoyi Dong, Yuhang Zang, and Jiaqi Wang.
\newblock Long-clip: Unlocking the long-text capability of clip, 2024.
\newblock URL \url{https://arxiv.org/abs/2403.15378}.

\bibitem[Zhu et~al.(2023)Zhu, Chen, Shen, Li, and Elhoseiny]{zhu2023minigpt4}
Deyao Zhu, Jun Chen, Xiaoqian Shen, Xiang Li, and Mohamed Elhoseiny.
\newblock Minigpt-4: Enhancing vision-language understanding with advanced large language models, 2023.

\end{thebibliography}
\bibliographystyle{arxiv}

\appendix
\section{Details of \Data}\label{sec:data_construction}
\begin{figure}[h!]
\includegraphics[width=1.0\textwidth]{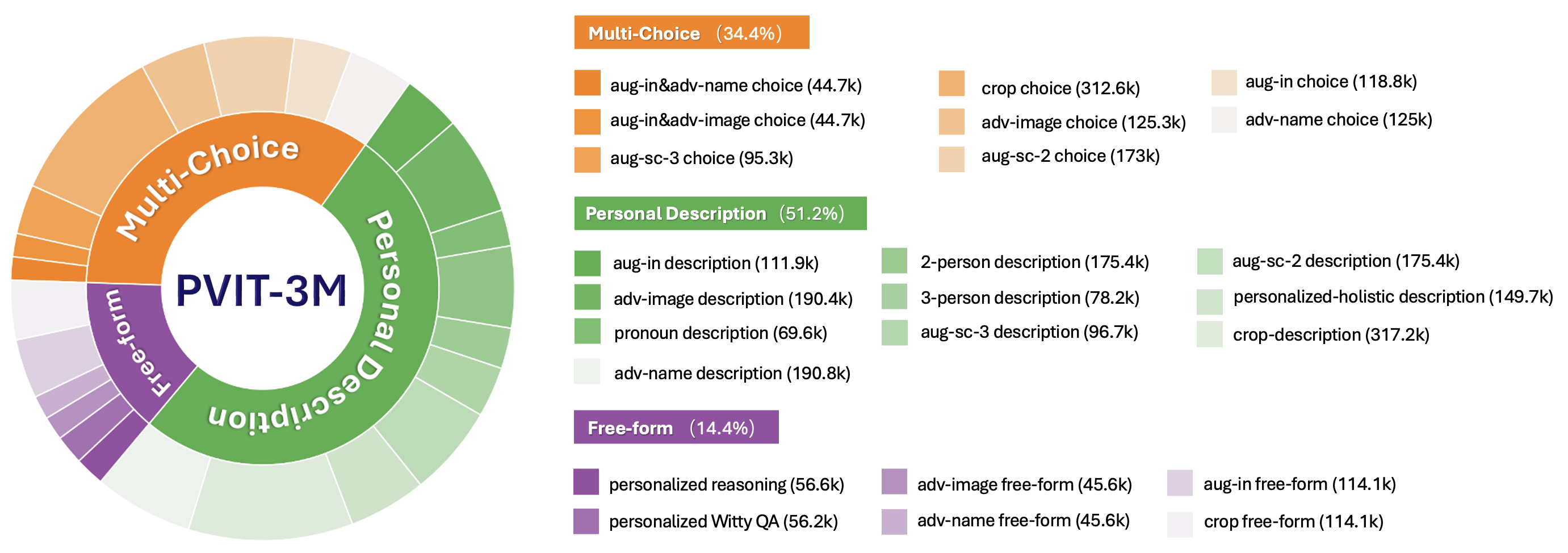} 
\vspace{-0.6cm}
\caption{ \textbf{Statistics of PVIT-3M}, a large scale personalized instruct tuning dataset. Left: Data Distribution within Each Category. The outer circle shows the distribution of all data categories and the inner circle shows the distribution of data subsets. Right: The detailed quantities of datasets.
}\label{fig:stats}
\end{figure}

We elaborate the detailed statistics of our constructed \Data in Figure~\ref{fig:stats}. In addition, we provide explanations of different categories in Table~\ref{tab:dataset_categories}.

\begin{table}[h!]
\centering
\begin{tabular}{c|p{0.60\textwidth}}
\Xhline{3\arrayrulewidth} 
\textbf{Keyword} & \textbf{Explanation} \\
\Xhline{3\arrayrulewidth} 
\multirow{1}{*}{Crop} & Uses cropped faces in the personalized prefix. \\
\hline
\multirow{2}{*}{Aug-In} & Uses generated photos based on the cropped faces in the personalized prefix. \\
\hline
\multirow{4}{*}{Aug-Sc-x} & Involves concatenating different cropped individuals into one image to replace the original scene image. The number ``x" following ``Aug-Sc" indicates the number of individuals concatenated. \\
\hline
\multirow{2}{*}{Adv-Name} & Indicates an unanswerable scenario where the name in the prefix is different from the name mentioned in the QA. \\
\hline
\multirow{2}{*}{Adv-Image} & Indicates an unanswerable scenario where the individual photo in the prefix is not present in the scene image. \\
\hline
\multirow{1}{*}{Pronoun Description} & Replaces specific names with pronouns in the description. \\
\hline
\multirow{3}{*}{x-person description} & Asks the MLLM to give a description of ``x" individuals while providing augmented scene image that concats ``x" persons. \\
\hline
\multirow{4}{*}{Personalized-Holistic Description} & The query is changed to a general one, such as ``describe this image", while requiring the model to include certain individual's name from the prefix in the overall description of the image. \\
\hline
\multirow{2}{*}{Personalized Reasoning} & Asks reasoning-based questions involving the specific individual. \\
\hline
\multirow{2}{*}{Personalized Witty QA} & Generates humorous and creative responses involving the specific individual. \\
\Xhline{3\arrayrulewidth}
\end{tabular}
\label{tab:dataset_categories}
\caption{Explanation of PVIT-3M Dataset Categories.}
\end{table}

\section{Details of \Bench}\label{sec:bench_stats}
\paragraph{Benchmark Data Curation} Directly annotating the data manually introduces tedious human labour. To boost the annotation efficiency, we adopt an LLM-assisted annotation pipeline: we follow a similar annotation procedure as the one adopted in LLaMA3~\citep{dubey2024llama3herdmodels}. Specifically, we first feed the images containing people to GPT4o, and instruct it to design QAs for each individual. Then, we manually check the validity of the designed QAs, and match them to the corresponding individuals.

\paragraph{Benchmark Statistics}
P-Bench contains two main categories: Multiple Choice (MC) and Description. The statistics for each category are detailed below, along with the total counts.

\subsection*{Multiple Choice (MC) Category}
\begin{itemize}
    \item \textbf{Type Statistics:} The MC category includes six types of samples, totaling 915 samples:
    \begin{itemize}
        \item Aug-Sc-2: 100 samples
        \item Aug-Sc-3: 100 samples
        \item Adv-Image: 100 samples
        \item Adv-Name: 100 samples
        \item Aug: 100 samples
        \item Crop: 415 samples
    \end{itemize}
    \item \textbf{Number of People in Scene Image:} Total of 415 images:
    \begin{itemize}
        \item Images containing 1 person: 221
        \item Images containing 2 people: 85
        \item Images containing 3 people: 48
        \item Images containing 4 or more people: 61
    \end{itemize}
\end{itemize}

\subsection*{Description Category}
\begin{itemize}
    \item \textbf{Type Statistics:} The Description category includes a total of 100 samples, divided as follows:
    \begin{itemize}
        \item \textbf{Answerable (Aug-In):} 60 samples
        \item \textbf{Unanswerable:} 40 samples, consisting of:
        \begin{itemize}
            \item Adv-Image: 20 samples
            \item Adv-Name: 20 samples
        \end{itemize}
    \end{itemize}
    \item \textbf{Number of People in Scene Image:} Total of 60 images:
    \begin{itemize}
        \item Images containing 1 person: 14
        \item Images containing 2 people: 31
        \item Images containing 3 people: 6
        \item Images containing 4 or more people: 9
    \end{itemize}
\end{itemize}


\begin{table*}[h!]
\centering
\begin{minipage}{1.0\textwidth}
\vspace{0mm}    
\centering
\begin{sectionbox}[]{Holistic Information Generation Prompt}
    \centering
      \footnotesize
    \begin{tabular}{p{0.97\textwidth} c}
$<$image$>$\\
Provide a detailed description of this image, with special emphasis on the main character, including their appearance, expressions, actions, and any distinguishing features.
    \end{tabular}
\end{sectionbox}
\vspace{-2mm}
\caption{The prompt for generating \textit{holistic information}.}
\label{tab:holistic info prompt}
\end{minipage}
\end{table*}

\begin{table*}[h!]
\centering
\begin{minipage}{1.0\textwidth}
\vspace{0mm}    
\centering
\begin{sectionbox}[]{Personal Information Generation Prompt}
    \centering
      \footnotesize
    \begin{tabular}{p{0.97\textwidth} c}
$<$image$>$\\
Describe the person in this image. Focus on this person's main features. Remember, **Do Not** include any background information. Additionally, in your response, you should use $<$name$>$ to refer to the person you describe when you mention the person's name first time. Again, you must contain $<$name$>$ in your response.
    \end{tabular}
\end{sectionbox}
\vspace{-2mm}
\caption{The prompt for generating specific \textit{person information}. $<$image$>$ represents the cropped photo of individual from the scene image. Here, we ask the MLLM to use the placeholder $<$name$>$ to refer the individual when describing.}
\label{tab:person info prompt}
\end{minipage}
\end{table*}

\section{Detailed Prompts for Data Generation}\label{sec:prompts}
We provide the detailed prompts for each phase of data generation. In Table~\ref{tab:holistic info prompt} and Table~\ref{tab:person info prompt}, we showcase the prompts provided to the MLLM, which are used for generating the holistic information and the personal information, respectively. In Table~\ref{tab:replace_person_prompt}, we demonstrate the prompt used for integrating the dual-level information into a single description, which is provided to the LLM along with the holistic and personalized information. In Table~\ref{tab:mc_prompt}, we demonstrate the prompt to LLM that is used for creating the Multi-Choice QA. For dual-level information fusion and  Multi-Choice QA generation, we incorporate manually designed in-context examples to guide the LLM.
\begin{table*}[t!]
\centering
\begin{minipage}{1.0\textwidth}
\vspace{0mm}    
\centering
\begin{sectionbox}[]{Dual-level Information Fusion Prompt}
    \centering
      \footnotesize
    \begin{tabular}{p{0.97\textwidth} c}
\textbf{\# TASK DESCRIPTION }\\
\vspace{1pt}
Given the Person Information of a person and the Holistic Information of the whole image, you need to put the placeholder \textless name \textgreater in the Holistic Information to represent the person described in Person Information.. \\\\

\textbf{\# EXAMPLE}\\
\vspace{1pt}
\textbf{Person Information}: \textit{$<$name$>$ is sitting in a relaxed posture. $<$name$>$ is wearing a dark blue T-shirt and light blue jeans. On his left wrist, he has a watch. $<$name$>$ is smiling and appears to be in a jovial mood. He is holding a blue object in his right hand, which looks like a piece of cloth or a towel. The background is a simple blue wall, and there is a light-colored blanket or couch behind him.}\\
\vspace{3pt}
\textbf{Holistic Information}: \textit{The image captures a heartwarming moment between a man and a young boy. The man, wearing glasses and a blue shirt, is sitting on a couch, holding a blue balloon in his hand. He is smiling and looking at the boy, who is seated next to him. The boy, wearing a green shirt, is also smiling and looking at the man. The background of the image features a blue wall, adding to the overall warmth of the scene. The man's position on the left and the boy's on the right create a balanced composition. The blue balloon held by the man adds a playful element to the image. The smiles on their faces suggest a moment of joy and connection between the two.}\\
\vspace{3pt}
\textbf{Integrated Information}: \textit{The image captures a heartwarming moment between $<$name$>$ and a young boy. \textless name\textgreater, wearing glasses and a blue shirt, is sitting on a couch, holding a blue balloon in his hand. He is smiling and looking at the boy, who is seated next to him. The boy, wearing a green shirt, is also smiling and looking at \textless name\textgreater. The background of the image features a blue wall, adding to the overall warmth of the scene. \textless name\textgreater's position on the left and the boy's on the right create a balanced composition. The blue balloon held by $<$name$>$ adds a playful element to the image. The smiles on their faces suggest a moment of joy and connection between the two.}\\\\

\vspace{1pt}
\textbf{\# TASK}\\
Person Information:  \{Person Information\}\\
Holistic Information: \{Holistic Information\}\\
Integrated Information: \\
    \end{tabular}
\end{sectionbox}
\vspace{-2mm}
\caption{The prompt for fusing person information and  holistic information. The \{Person Information\} and \{Holistic Information\} are placeholders that will be replaced by the actual content to be processed.}
\label{tab:replace_person_prompt}
\end{minipage}
\end{table*}

\begin{table*}[h!]
\centering
\begin{minipage}{1.0\textwidth}
\vspace{0mm}    
\centering
\begin{sectionbox}[]{Multi-Choice QA Generation Prompt}
    \centering
      \footnotesize
    \begin{tabular}{p{0.97\textwidth} c}
\textbf{\# TASK DESCRIPTION}\\
\vspace{1pt}
Now you need to generate multiple-choice questions based on Information. You should pay particular attention to the characteristics mentioned in the description that describe this person, and use these characteristics to create questions and possible answers. \\\\
\textbf{\# RESPONSE FORMAT}\\
\vspace{1pt}
Your response must strictly follow the format below:

[[{``question": ``…", ``choices": [``…", ``…", ``…", ``…"], ``answer": ``…"}]]\\
\\
\textbf{\# ATTENTION}\\
\vspace{1pt}
1. Please ensure that all references to the person in your questions and answers are replaced with the placeholder \textless name \textgreater. \\
2. Only generate multiple-choice questions about the individual.\\
3. Ensure that each set of choices has clear distinctions and no overlap to avoid multiple correct answers.\\
\vspace{3pt}
\textbf{\# EXAMPLE}\\
\vspace{1pt}
\textbf{Information}: \textit{In the photo, $<$name$>$ is wearing a white shirt and blue jeans. She is standing beside a man in a blue T-shirt and has her hands on her hips. She is also wearing a black bag.}\\\\

\textbf{Generated MC}: \textit{[{``question": "What color shirt is $<$name$>$  wearing?", ``choices": [``Red", ``White", ``Blue", ``Black"], ``answer": ``White"}], [{``question": ``What color are $<$name$>$'s jeans?", ``choices": [``Black", ``Green", ``Blue", ``Yellow"], ``answer": ``Blue"}], [{``question": ``What is $<$name$>$ doing with her hands?", ``choices": [``Holding a bag", ``Hands on her hips", ``Waving", ``In her pockets"], ``answer": ``Hands on her hips"}], [{``question": ``What accessory is $<$name$>$ wearing?", ``choices": [``A hat", ``A scarf", ``A black bag", ``Sunglasses"], ``answer": ``A black bag"}]}
\vspace{4pt}
\\\\
\textbf{\#TASK}\\
\vspace{1pt}
Information: \{Information\}\\\\

Generated MC: 
    \end{tabular}
\end{sectionbox}
\vspace{-2mm}
\caption{Prompt for generating Multi-Choice QA.}
\label{tab:mc_prompt}
\end{minipage}
\end{table*}

\begin{table*}[h!]
\centering
\caption{Hyper-parameters used for \ModelNameAbbre.}
\label{tab:deepspeed_params}
\resizebox{0.75\textwidth}{!}{
\begin{tabular}{l|l}
\toprule
\textbf{Parameter} & \textbf{Value} \\
\midrule
\texttt{--lora\_enable} & \texttt{True} \\
\texttt{--lora\_r} & \texttt{128} \\
\texttt{--lora\_alpha} & \texttt{256} \\
\texttt{--mm\_projector\_lr} & \texttt{1e-4} \\
\texttt{--deepspeed} & \texttt{./scripts/zero2.json} \\
\texttt{--version} & \texttt{v1} \\
\texttt{--vision\_tower} & \texttt{openai/clip-vit-large-patch14-336} \\
\texttt{--mm\_projector\_type} & \texttt{mlp2x\_gelu} \\
\texttt{--mm\_vision\_select\_layer} & \texttt{-2} \\
\texttt{--mm\_use\_im\_start\_end} & \texttt{False} \\
\texttt{--mm\_use\_im\_patch\_token} & \texttt{False} \\
\texttt{--image\_aspect\_ratio} & \texttt{pad} \\
\texttt{--group\_by\_modality\_length} & \texttt{True} \\
\texttt{--bf16} & \texttt{True} \\
\texttt{--num\_train\_epochs} & \texttt{1} \\
\texttt{--per\_device\_train\_batch\_size} & \texttt{16} \\
\texttt{--per\_device\_eval\_batch\_size} & \texttt{4} \\
\texttt{--gradient\_accumulation\_steps} & \texttt{2} \\
\texttt{--evaluation\_strategy} & \texttt{no} \\
\texttt{--save\_strategy} & \texttt{steps} \\
\texttt{--save\_steps} & \texttt{50000} \\
\texttt{--save\_total\_limit} & \texttt{1} \\
\texttt{--learning\_rate} & \texttt{2e-4} \\
\texttt{--weight\_decay} & \texttt{0.} \\
\texttt{--warmup\_ratio} & \texttt{0.03} \\
\texttt{--lr\_scheduler\_type} & \texttt{cosine} \\
\texttt{--logging\_steps} & \texttt{1} \\
\texttt{--tf32} & \texttt{True} \\
\texttt{--model\_max\_length} & \texttt{4096} \\
\texttt{--gradient\_checkpointing} & \texttt{True} \\
\texttt{--dataloader\_num\_workers} & \texttt{4} \\
\texttt{--lazy\_preprocess} & \texttt{True} \\
\bottomrule
\end{tabular}
}
\end{table*}

\section{Training Details}\label{sec:training_detail}
In Table~\ref{tab:deepspeed_params}, we illustrate the detailed hyper-parameters used when fine-tuning the MLLM with our \Data. We wish to note that we start tuning from the checkpoint of LLaVA-7B~\cite{liu2023llava}. We train the MLLM with a subset of our \Data with 1M samples.  The entire training is conducted on 8 A100 GPUs with 80GB memory, which lasted for 30 hours.

\end{document}